\documentclass[sigconf,nonacm]{acmart}
\AtBeginDocument{%
  }

\setcopyright{none}

\usepackage{amsmath}
\usepackage{multirow}
\usepackage{placeins}

\begin{document}
\setlength{\textfloatsep}{6pt}
\setlength{\intextsep}{6pt}
\setlength{\floatsep}{6pt}
\setlength{\dbltextfloatsep}{6pt}
\setlength{\dblfloatsep}{6pt}
\captionsetup[figure]{skip=4pt,belowskip=0pt}
\raggedbottom
\emergencystretch=3em
\title[When Fusion Order Matters]{When Fusion Order Matters: Insights on Progressive Multimodal Representation Learning for Cross-modal Retrieval}
\author{\texorpdfstring{\normalsize Haoxiang Zhao\textsuperscript{1,2}, Bin Wang\textsuperscript{2}, Yaofei Wang\textsuperscript{2}, Xiao Lan\textsuperscript{3}, Yunfeng Diao\textsuperscript{2}, Xu Wang\textsuperscript{1}}{Haoxiang Zhao, Bin Wang, Yaofei Wang, Xiao Lan, Yunfeng Diao, Xu Wang}}
\affiliation{%
  \institution{\textsuperscript{1}University of Science and Technology of China\\
  \textsuperscript{2}Hefei University of Technology\quad
  \textsuperscript{3}Sichuan University}
  \city{}\country{}}
\renewcommand{\shortauthors}{Zhao et al.}
\begin{abstract}
Multimodal video retrieval combines visual, acoustic, and speech information, yet how their fusion order should depend on the data and the retrieval query remains unclear. We introduce Unified Progressive Representation Learning for Order Analysis, or UniPRO, to investigate this question through configurable fusion schedules with matched model components and training settings. Starting from text-to-video retrieval, we find that the tested one-by-one schedules outperform stacked fusion, while changing the order within one-by-one fusion also changes performance. To guide order selection, we examine two complementary considerations: learning difficulty under unequal modality availability and preservation of information relevant to the query. Complete-modality controls show that missing inputs alone cannot explain order sensitivity. At R@1, text and audio retrieval favor schedules that introduce the modality matching the query last. Controlled speech missingness during training and testing reverses the preferred order in both R@1 and SumR, with different crossover points. These findings motivate balancing the two considerations when selecting a fusion path. We then use them to predict a visual-last schedule for image queries, which achieves 97.9 R@1, the highest among the evaluated orders on MSR-VTT-9k. Main experiments on four text-to-video benchmarks establish the effectiveness of UniPRO, with component ablations assessing its supporting design. Together, the results identify fusion order as a task-dependent choice that should account jointly for modality availability and the retrieval query.
\end{abstract}
{\emergencystretch=3em\maketitle}
\hypersetup{pdfauthor={Haoxiang Zhao, Bin Wang, Yaofei Wang, Xiao Lan, Yunfeng Diao, Xu Wang}}
\begin{figure}[t]
\centering
\includegraphics[width=\columnwidth]{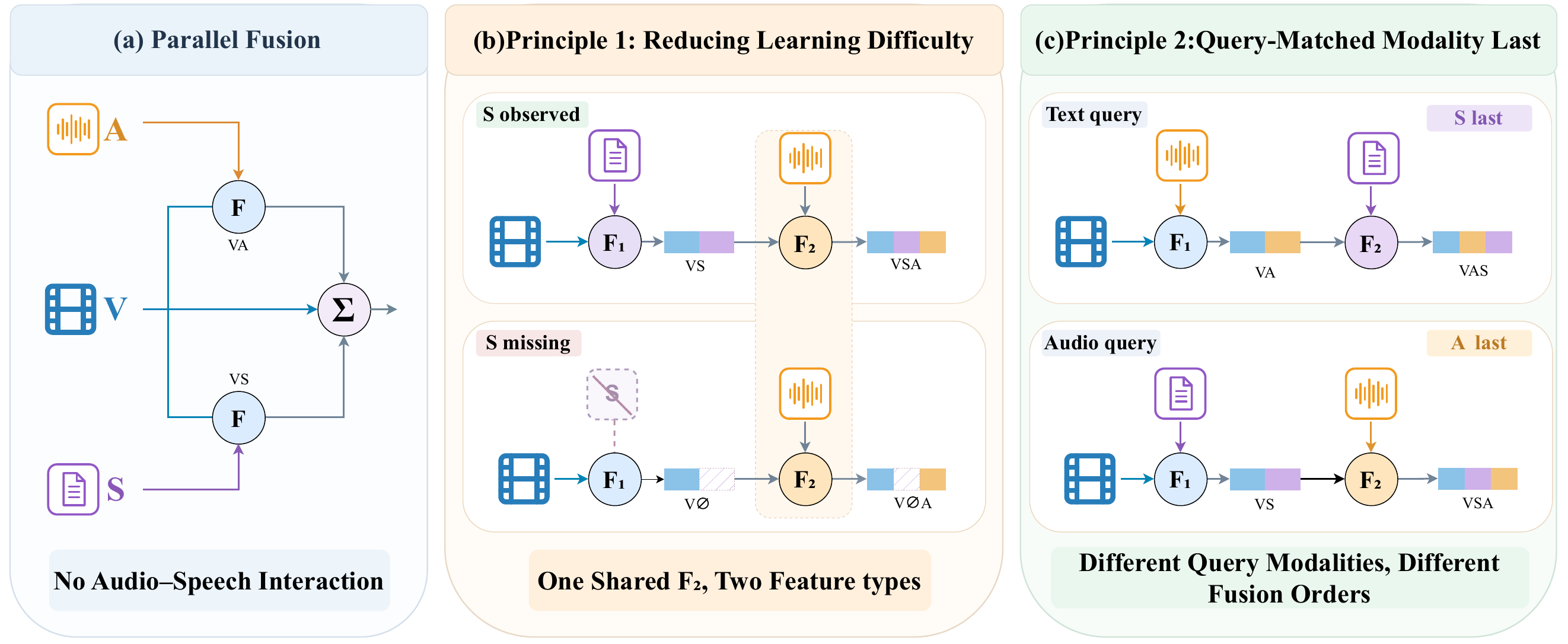}
\caption{Fusion order matters: unordered fusion, an unsuitable stacked order, and a preferred one-by-one order for text queries.}
\Description{Three panels contrast unordered fusion, audio and speech fused before vision, and audiovisual fusion followed by speech semantics.}
\label{fig:order-teaser}
\end{figure}

\section{Introduction}
\label{sec:introduction}
{\emergencystretch=3em
Short-video platforms such as Kuaishou and TikTok contain rich visual, acoustic, and spoken information. Effective retrieval requires combining these complementary cues into a representation suitable for matching a query. The fusion process determines how the cues interact and contribute to matching, making its design central to multimodal video representation learning.\par}

{\emergencystretch=3em
Recent advances have improved retrieval through stronger representations and richer modality coverage. CLIP-based methods adapt pretrained visual and language features to video retrieval~\cite{radford2021clip,luo2022clip4clip,gorti2022xpool}; audiovisual methods incorporate acoustic cues~\cite{lin2022eclipse,ibrahimi2023tefal,jeong2025avigate}; SAVE further separates speech semantics from general audio features~\cite{zhao2026save}. These approaches improve the available information and its alignment, but the same inputs do not specify the order in which their representations should interact. We ask how fusion order affects retrieval with the supporting components held fixed, and what can guide its selection.\par}

{\emergencystretch=3em
In progressive fusion, each interaction updates the representation passed to the next stage. An early modality can change the correspondences that a later stage must learn, while later interactions affect how earlier information contributes to matching. We introduce \emph{Unified Progressive Representation Learning for Order Analysis}, or UniPRO, to compare these paths within a common framework. UniPRO prepares and aligns modality features before applying configurable gated fusion stages. With matched encoders, components, and training settings, we compare stacked and one-by-one fusion, then vary the order within the latter. Figure~\ref{fig:order-teaser} illustrates these choices. The tested one-by-one schedules outperform stacked fusion, but their internal order also matters: on MSR-VTT-9k, introducing audio before speech yields 53.9 text-to-video R@1, versus 53.2 for the reverse order.\par}

{\emergencystretch=3em
Our first explanation concerns learning difficulty under unequal modality availability. Every sample has visual input, but usable audio and transcribed speech are less consistently available. A fusion stage may encounter different effective input patterns depending on the preceding order, motivating a path that simplifies the correspondences to be learned. To examine whether missingness explains the observed preference, we train and evaluate on complete samples with all three modalities. The order difference persists. Frozen-model tests further show that this learned difference can remain when inputs are removed only at inference. Missing inputs therefore cannot fully explain order sensitivity.\par}

{\emergencystretch=3em
This result motivates a second perspective: preserving information relevant to the query. Speech is introduced last in the preferred text retrieval schedule, suggesting that later fusion may help retain cues compatible with the query modality. Natural modality groups motivate this hypothesis. We examine its prediction for audio queries using complete MSR-VTT samples: introducing audio last achieves higher R@1 in both retrieval directions, although the advantage does not extend to every recall cutoff. These findings suggest complementary principles: favor a path suited to the observed input patterns, and favor later fusion of information matching the query.\par}

{\emergencystretch=3em
The principles need to be considered together. In audio retrieval, increasing speech missingness in both training and test sets, with models retrained at each rate, reverses the preferred order in both audio-to-video R@1 and video-to-audio SumR. The crossover depends on the metric. We use this reasoning to predict an order for image queries: integrate visual, acoustic, and speech information, then revisit the original visual representation last. It achieves 97.9 R@1 on the full MSR-VTT-9k split, the highest among the evaluated alternatives. This transfer supports using the insights to guide order selection for another query modality.\par}

{\emergencystretch=3em
Main experiments on four text-to-video benchmarks, reverse retrieval, and a second visual backbone assess the effectiveness of the complete UniPRO pipeline. Component ablations establish the contributions of transcript condensation, relation enhancement, and feature alignment. The order analyses address the separate question of how to choose the fusion path within that pipeline. Our contributions are threefold:\par}
\begin{enumerate}
\item We introduce UniPRO, a common framework for comparing fusion organization and modality order with matched representation components.
\item We identify learning difficulty and preservation of query-relevant information as complementary considerations for order selection, supported by complete-modality controls, query changes, and input-removal analyses.
\item We evaluate UniPRO across four text-to-video benchmarks and use the resulting principles to predict an effective fusion order for image queries.
\end{enumerate}
\section{Related Work}
\label{sec:related-work}

\subsection{Cross-Modal Video Retrieval}
\label{sec:cross-modal-video-retrieval}
{\emergencystretch=3em
Cross-modal video retrieval matches queries from different modalities to videos, with text-to-video retrieval receiving the most attention. CLIP provides a strong foundation for visual and language alignment~\cite{radford2021clip}. CLIP4Clip adapts this foundation to video retrieval through temporal aggregation and cross-modal matching~\cite{luo2022clip4clip}. X-Pool uses text-conditioned attention to aggregate relevant frames~\cite{gorti2022xpool}, while UCoFiA models visual and textual correspondences at multiple granularities~\cite{wang2023ucofia}. These methods primarily improve the match between text queries and visual content.\par}

{\emergencystretch=3em
Audio and image queries extend retrieval beyond language. MAViL combines masked reconstruction, contrastive learning, and self-training to learn audiovisual representations~\cite{huang2023mavil}; AVSiam studies a shared Transformer backbone for audio and vision~\cite{lin2024avsiam}. ImageBind and LanguageBind connect modalities through alignment centered on images and language, respectively~\cite{girdhar2023imagebind,zhu2024languagebind}. These approaches expand the available representations and query types. We examine how the order of candidate modality fusion affects retrieval across these query types.\par}

\subsection{Multimodal Video Representation Learning}
\label{sec:multimodal-video-representation}
{\emergencystretch=3em
Video provides complementary visual, acoustic, and spoken evidence. ECLIPSE combines sight and sound for efficient retrieval over long video sequences~\cite{lin2022eclipse}. TEFAL aligns text separately with visual and audio features to capture relevant cues~\cite{ibrahimi2023tefal}. AVIGATE regulates the contribution of audio through gated attention, suppressing irrelevant sounds while keeping query and video encoding independent~\cite{jeong2025avigate}.\par}

{\emergencystretch=3em
SAVE adds a separate speech-semantic branch and improves audiovisual alignment with soft-ALBEF~\cite{zhao2026save}. Together, these methods improve information coverage, relevance, and alignment. UniPRO builds on this setting by examining a further choice: how different fusion paths combine the same inputs and interaction components.\par}

\subsection{Multimodal Interaction and Progressive Fusion}
\label{sec:interaction-progressive-fusion}
{\emergencystretch=3em
Multimodal fusion depends on the direction and location of information exchange. MulT uses directional attention between modality pairs to model dependencies in unaligned sequences~\cite{tsai2019mult}. Multimodal Bottleneck Transformer restricts communication through bottleneck tokens at multiple layers~\cite{nagrani2021mbt}. These architectures establish interaction direction and communication paths as important design choices.\par}

{\emergencystretch=3em
Progressive fusion further updates representations through successive interactions. Pro-Fusion feeds information from late fusion back to earlier unimodal feature generators~\cite{shankar2022profusion}. Refinement across network depth differs from changing the order of modality inputs within a fixed fusion structure. UniPRO studies the latter by comparing stacked and one-by-one fusion, analyzing alternative orders, and testing whether the resulting insights extend across query modalities.\par}

\section{Method: UniPRO}
\label{sec:method}
\begin{figure*}[t]
\centering
\includegraphics[width=\textwidth]{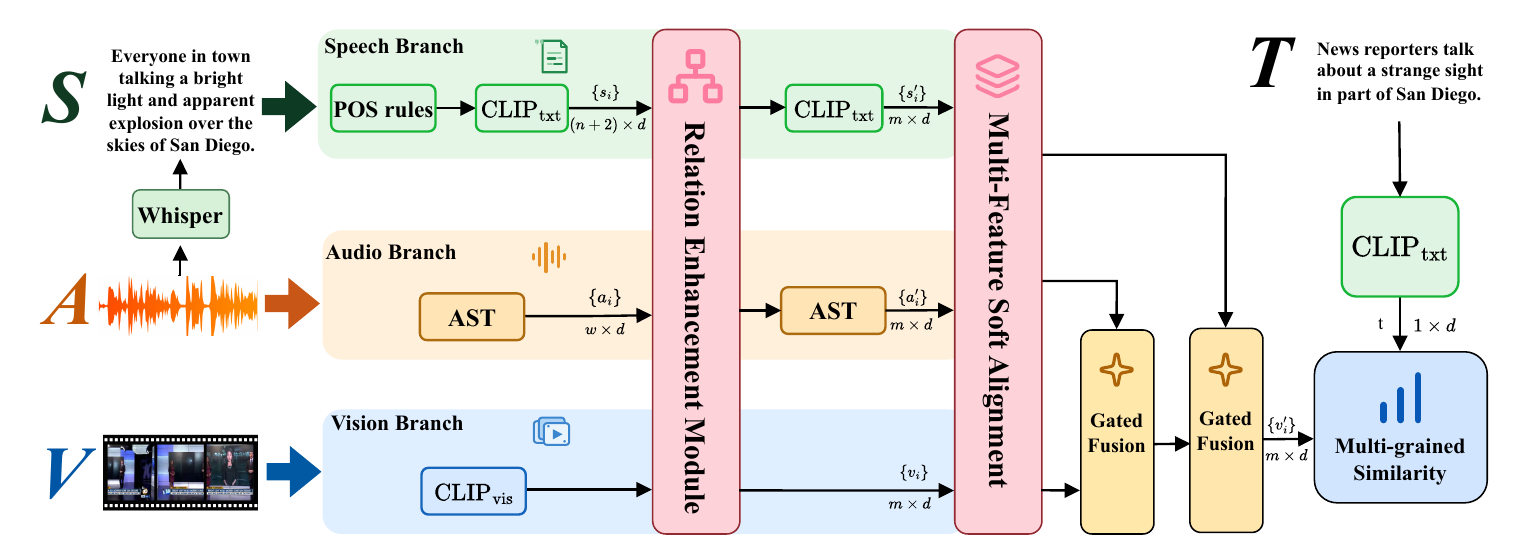}
\caption{UniPRO instantiated for text-to-video retrieval. POSR condenses speech transcripts; REM enhances audio--speech relations; MFSA aligns the three feature spaces. Two gated stages first condition visual features on audio and then on speech semantics. The final representation is compared with the independently encoded query.}
\Description{Visual, auditory, and transcribed speech features pass through relation enhancement and soft alignment before successive gated fusion with visual tokens as the residual stream.}
\label{fig:method-overview}
\end{figure*}

UniPRO constructs a multimodal candidate representation from video frames, audio, and available speech transcripts. Figure~\ref{fig:method-overview} shows its text-to-video instantiation. We use V, A, and S for visual, acoustic, and speech-semantic features, respectively. T denotes the retrieval text query and is distinct from the transcript represented by S. The pipeline first prepares and aligns the modality features, then combines them according to a specified fusion schedule.

\subsection{Modality Representations}
\label{sec:preliminaries-motivation}

{\emergencystretch=3em
We follow the modality encoding pipeline of SAVE~\cite{zhao2026save}, summarized in Equations~\ref{eq:save-visual-audio-tokens} and~\ref{eq:save-speech-tokens}. For a video \(\mathcal{V}\), we uniformly sample \(m\) frames. The CLIP visual encoder, \(\operatorname{CLIP}_{\mathrm{vis}}\), represents each frame by its \(d\)-dimensional \([\mathrm{CLS}]\) token from the final encoder layer~\cite{radford2021clip}. This yields the visual sequence \(\{v_i\}_{i=1}^{m}\).\par}

\begin{equation}
\left\{
\begin{aligned}
\{f_1,\ldots,f_m\}
&\leftarrow \operatorname{video\text{-}to\text{-}frames}(\mathcal{V},m),\\
\{v_1,\ldots,v_m\}
&\leftarrow \operatorname{CLIP}_{\mathrm{vis}}(\{f_1,\ldots,f_m\}),\\
\{a_1,\ldots,a_m\}
&\leftarrow \operatorname{Resampler}(\operatorname{AST}(\mathcal{A}),m).
\end{aligned}
\right.
\label{eq:save-visual-audio-tokens}
\end{equation}

{\emergencystretch=3em
The source soundtrack \(\mathcal{A}\) is converted to Mel filter-bank features and encoded by AST~\cite{gong2021ast}. A resampler maps the resulting acoustic sequence to \(m\) tokens. For speech semantics, Whisper large-v3 transcribes the soundtrack into words \(\{w_1,\ldots,w_n\}\)~\cite{radford2023whisper}. The CLIP text encoder then produces \(d\)-dimensional tokens \(\{s_S,s_1,\ldots,s_n,s_E\}\), including the start and end tokens \(s_S\) and \(s_E\)~\cite{radford2021clip}.\par}

\begin{equation}
\left\{
\begin{aligned}
\{w_1,\ldots,w_n\}
&\leftarrow \operatorname{Whisper}(\mathcal{A}),\\
\{s_S,s_1,\ldots,s_n,s_E\}
&\leftarrow \operatorname{CLIP}_{\mathrm{txt}}(\{w_1,\ldots,w_n\}).
\end{aligned}
\right.
\label{eq:save-speech-tokens}
\end{equation}

{\emergencystretch=3em
Following AVIGATE~\cite{jeong2025avigate}, we encode the retrieval query \(T\) with \(\operatorname{CLIP}_{\mathrm{txt}}\) to obtain \(t\). Global similarity \(s_g(V,T)\) measures the cosine similarity between the video representation and \(t\). Local similarity \(s_l(V,T)\) aggregates the cosine similarities between individual fused visual tokens and \(t\) through logarithmic and exponential operations. The final matching score averages these two similarities as \((s_g+s_l)/2\).\par}

\subsection{Component Design}
\label{sec:component-design}

\subsubsection{POS Rules}
\label{sec:posr}

{\emergencystretch=3em
ASR transcripts can be lengthy and noisy, particularly in complex speech environments such as e-commerce livestreams. Direct truncation to the CLIP input limit may discard useful semantic cues. We therefore introduce part-of-speech rules, abbreviated POSR, to condense transcripts according to word category. It favors words likely to carry the entities, actions, and attributes needed for retrieval while adding little computational overhead.\par}

{\emergencystretch=3em
We tokenize the transcript with spaCy and assign Universal Dependencies UPOS tags~\cite{nivre2020ud}. Equation~\ref{eq:pos-priority} defines the retention priorities. Words from higher-priority categories are selected first; uniform sampling resolves selection within the same category. The condensed transcript respects the CLIP input budget, retaining semantic coverage instead of relying only on a fixed prefix.\par}

\begin{equation}
P(w_i)=
\begin{cases}
4, & \operatorname{POS}(w_i)\in\{\mathrm{PROPN},\mathrm{NOUN}\},\\
3, & \operatorname{POS}(w_i)=\mathrm{VERB},\\
2, & \operatorname{POS}(w_i)=\mathrm{ADJ},\\
1, & \operatorname{POS}(w_i)\in\{\mathrm{ADV},\mathrm{NUM}\}.
\end{cases}
\label{eq:pos-priority}
\end{equation}

\subsubsection{Relation Enhancement Module}
\label{sec:rem}

{\emergencystretch=3em
Acoustic features and speech semantics describe different aspects of the same soundtrack. Encoding them separately can leave useful relations implicit. The relation enhancement module, REM, strengthens these relations through bidirectional attention before progressive fusion, allowing each stream to incorporate evidence from the other.\par}

{\emergencystretch=3em
Let \(S=\{s_i\}_{i=1}^{L_s}\) and \(A=\{a_i\}_{i=1}^{L_a}\) denote the speech and audio sequences. A Transformer resampler with \(x\) learnable queries compresses each stream, as shown in Equation~\ref{eq:rem-resampling}. Each original sequence then attends to the compressed sequence from the other modality. The attention update preserves the original sequence dimensions and is added through a residual connection. Finally, the resampler from SAVE produces \(m\) tokens for each enhanced stream, yielding \(\mathbf{S}'=\{s_i'\}_{i=1}^{m}\) and \(\mathbf{A}'=\{a_i'\}_{i=1}^{m}\) in Equation~\ref{eq:rem-enhancement}.\par}

\begin{equation}
\bar{\mathbf{S}}
=
\mathcal{R}_{x}(\mathbf{S}),
\qquad
\bar{\mathbf{A}}
=
\mathcal{R}_{x}(\mathbf{A})
\label{eq:rem-resampling}
\end{equation}

\begin{equation}
\begin{aligned}
\mathbf{S}'
&=
\mathcal{R}_{m}\left(
\mathbf{S}
+
\operatorname{Attn}
\left(
\mathbf{S},
\bar{\mathbf{A}},
\bar{\mathbf{A}}
\right)
\right),\\
\mathbf{A}'
&=
\mathcal{R}_{m}\left(
\mathbf{A}
+
\operatorname{Attn}
\left(
\mathbf{A},
\bar{\mathbf{S}},
\bar{\mathbf{S}}
\right)
\right).
\end{aligned}
\label{eq:rem-enhancement}
\end{equation}

\subsubsection{Multi-Feature Soft Alignment}
\label{sec:mfsa}

{\emergencystretch=3em
The gated module in Equation~\ref{eq:gated-fusion} combines a residual visual stream with an auxiliary modality. Its attention weights depend on the compatibility of the query and key features, motivating alignment before fusion. SAVE uses soft-ALBEF~\cite{zhao2026save} to address noisy audiovisual pairs when adapting ALBEF~\cite{li2021albef}. Its supervision is derived from ImageBind~\cite{girdhar2023imagebind}, whose embedding geometry differs from the shared CLIP space of visual and speech features. Aligning only vision and audio therefore leaves the relations involving speech insufficiently constrained. We extend this approach with multi-feature soft alignment, MFSA, to supervise all three modality pairs. Section~\ref{sec:progressive-fusion} describes how the aligned streams enter successive fusion stages.\par}

\begin{equation}
\begin{aligned}
\operatorname{GFM}(\mathbf{V},\mathbf{X})
&=
\mathbf{V}
+
\mathbf{G}\odot
\operatorname{Attn}(\mathbf{V},\mathbf{X},\mathbf{X}),\\
\mathbf{G}
&=
\sigma\left(
\mathbf{W}_g
\left[
\mathbf{V};
\operatorname{Attn}(\mathbf{V},\mathbf{X},\mathbf{X})
\right]
+\mathbf{b}_g
\right).
\end{aligned}
\label{eq:gated-fusion}
\end{equation}

{\emergencystretch=3em
MFSA aligns visual and acoustic features, visual and speech features, and acoustic and speech features. For each pair \((p,q)\in\{(v,a),(v,s),(a,s)\}\), ImageBind embeddings define a reference affinity matrix \(\mathbf{M}_{0}^{p,q}\in\mathbb{R}^{B\times B}\) over a batch of \(B\) examples~\cite{girdhar2023imagebind}. Each entry is a cosine similarity between examples from the two modalities. We compute \(\mathbf{M}_{1}^{p,q}\) from the current network in the same way, then use the reference matrix as soft supervision for its row and column similarity distributions:\par}

\begin{equation}
\begin{aligned}
\mathcal{L}_{\mathrm{SA}}^{p,q}
&=
\frac{1}{B}\sum_{i=1}^{B}
d_p\left(
\sigma(\mathbf{M}_{0}^{p,q}[i,:]),
\sigma(\mathbf{M}_{1}^{p,q}[i,:])
\right)\\
&\quad+
\frac{1}{B}\sum_{j=1}^{B}
d_p\left(
\sigma(\mathbf{M}_{0}^{p,q}[:,j]),
\sigma(\mathbf{M}_{1}^{p,q}[:,j])
\right),
\end{aligned}
\label{eq:soft-alignment}
\end{equation}

{\emergencystretch=3em
where \(\sigma\) is softmax and \(d_p\) is the Pearson distance. Unlike losses based on absolute affinity values, such as MSE and Huber loss, Pearson distance emphasizes relative similarity patterns. This makes it suitable for preserving the ranking relations used in retrieval~\cite{huang2022dist,tian2024teachclip}.\par}

{\emergencystretch=3em
The overall MFSA loss averages the three pairwise terms:\par}

\begin{equation}
\mathcal{L}_{\mathrm{MFSA}}
=
\frac{1}{3}
\left(
\mathcal{L}_{\mathrm{SA}}^{v,a}
+
\mathcal{L}_{\mathrm{SA}}^{v,s}
+
\mathcal{L}_{\mathrm{SA}}^{a,s}
\right).
\label{eq:multi-feature-soft-alignment}
\end{equation}

\subsection{A Unified View of Progressive Fusion}
\label{sec:progressive-fusion}
Let \(\mathcal{F}_{\ell}(H,X)\) denote the gated fusion stage in Equation~\ref{eq:gated-fusion}. H supplies the attention query and residual stream, while X supplies the keys and values. A one-by-one schedule passes the updated representation from one stage to the next:
\begin{equation}
H^{(0)}=X_{\pi_1},\qquad
H^{(\ell)}=\mathcal{F}_{\ell}(H^{(\ell-1)},X_{\pi_{\ell+1}}).
\label{eq:unified-schedule}
\end{equation}
Here \(\pi\) specifies the order of modality streams. The text retrieval comparison starts from the visual stream and applies either audio followed by speech or speech followed by audio:
\begin{equation}
\begin{aligned}
H_{VAS}&=\mathcal{F}_2(\mathcal{F}_1(V,A),S),\\
H_{VSA}&=\mathcal{F}_2(\mathcal{F}_1(V,S),A).
\end{aligned}
\label{eq:vas-vsa}
\end{equation}
A schedule specifies both the order and the direction of interaction. The stacked ASV variants first combine audio and speech, then interact with vision. We distinguish the variants by whether A or S supplies the first attention query. This distinction is necessary because the modality list ASV alone does not specify the attention direction.

{\emergencystretch=3em
All schedules use the same feature extraction and matching components, allowing us to examine how the organization of fusion affects the use of available evidence. Passing an updated representation to a later stage changes the information and optimization path encountered by that stage. The comparison therefore tests the effectiveness of specific schedules rather than assuming that progressive fusion is always beneficial.\par}

\subsection{Training and Transfer Across Query Modalities}
Training combines the adaptive margin contrastive objective from AVIGATE~\cite{jeong2025avigate} and the MFSA loss with equal weights. MFSA also weights its three pairwise terms equally. Candidate representations are matched to independently encoded queries: CLIP encodes text and image queries, while the acoustic encoder handles audio queries. The candidate video retains the modalities available under the evaluation protocol.

For image queries, we add a third gated stage that revisits the original visual tokens. VASV and VSAV both end with V and differ in the preceding order of A and S. VVAS and VAVS instead introduce the repeated V earlier. These four schedules have three independently parameterized fusion stages, while VAS provides the original reference with two stages. Repeated V always denotes the original visual representation, with frame positions added once.
\begin{table*}[!t]
\caption{Text-to-video retrieval on four benchmarks.}
\label{tab:sota-comparison}
\centering
\small
\setlength{\tabcolsep}{1.2pt}
\begin{tabular}{@{}l*{16}{r}@{}}
\toprule
\multicolumn{1}{c}{\multirow{2}{*}{\textbf{Model}}}
& \multicolumn{4}{c}{\textbf{MSR-VTT-9k}}
& \multicolumn{4}{c}{\textbf{MSR-VTT-7k}}
& \multicolumn{4}{c}{\textbf{VATEX}}
& \multicolumn{4}{c}{\textbf{Charades}}\\
\cmidrule(lr){2-5}\cmidrule(lr){6-9}\cmidrule(lr){10-13}\cmidrule(lr){14-17}
& R@1 & R@5 & R@10 & SumR
& R@1 & R@5 & R@10 & SumR
& R@1 & R@5 & R@10 & SumR
& R@1 & R@5 & R@10 & SumR\\
\midrule
\multicolumn{17}{l}{\textit{Vision-only methods}}\\
CLIP4Clip, Neurocomputing'22
& 44.5 & 71.4 & 81.6 & 197.5
& 29.4 & 54.9 & 65.8 & 150.1
& 61.6 & 91.1 & 95.8 & 248.5
& 17.7 & 39.5 & 50.4 & 107.6\\
X-Pool, CVPR'22
& 46.9 & 72.8 & 82.2 & 201.9
& -- & -- & -- & --
& -- & -- & -- & --
& -- & -- & -- & --\\
TS2-Net, ECCV'22
& 47.0 & 74.5 & 83.8 & 205.3
& 29.9 & 56.4 & 67.3 & 153.6
& 59.1 & 90.0 & 95.2 & 244.3
& 16.9 & 38.5 & 49.5 & 104.9\\
X-CLIP, MM'22
& 46.1 & 73.0 & 83.1 & 202.2
& 31.2 & 57.4 & 68.1 & 156.7
& 62.2 & 90.9 & 95.4 & 248.5
& 18.9 & 41.3 & 53.0 & 113.2\\
CLIP-ViP, ICLR'23
& 46.5 & 72.1 & 82.5 & 201.1
& 30.2 & 56.0 & 69.1 & 155.3
& 62.1 & 90.2 & 97.0 & 249.3
& -- & -- & -- & --\\
STAN, CVPR'23
& 46.9 & 72.8 & 82.8 & 202.5
& -- & -- & -- & --
& -- & -- & -- & --
& -- & -- & -- & --\\
DiCoSA, IJCAI'23
& 47.5 & 74.7 & 83.8 & 206.0
& -- & -- & -- & --
& -- & -- & -- & --
& -- & -- & -- & --\\
PromptSwitch, ICCV'23
& 47.8 & 73.9 & 82.2 & 203.9
& -- & -- & -- & --
& -- & -- & -- & --
& -- & -- & -- & --\\
UATVR, ICCV'23
& 47.5 & 73.9 & 83.5 & 204.9
& -- & -- & -- & --
& 61.3 & 91.0 & 95.6 & 247.9
& -- & -- & -- & --\\
UCoFiA, ICCV'23
& 48.2 & 73.3 & 82.3 & 203.8
& 29.9 & 56.0 & 66.7 & 152.6
& 61.1 & 90.5 & -- & --
& 17.6 & 38.6 & 48.6 & 104.8\\
ProST, ICCV'23
& 48.2 & 74.6 & 83.4 & 206.2
& 30.4 & 56.7 & 67.7 & 154.8
& 60.6 & 90.5 & 95.4 & 246.5
& 17.8 & 37.0 & 48.3 & 103.1\\
DGL, AAAI'24
& 45.8 & 69.3 & 79.4 & 194.5
& -- & -- & -- & --
& 56.2 & 87.1 & 93.5 & 236.8
& -- & -- & -- & --\\
EERCF, AAAI'24
& 47.8 & 74.1 & 84.1 & 206.0
& 31.5 & 57.4 & 67.6 & 156.5
& 62.6 & 91.5 & 95.8 & 249.9
& -- & -- & -- & --\\
TeachCLIP, CVPR'24
& 46.8 & 74.3 & 82.6 & 203.7
& 30.9 & 57.1 & 68.0 & 156.0
& 63.6 & 91.9 & 96.1 & 251.6
& 19.2 & 40.1 & 51.2 & 110.5\\
TempMe, ICLR'25
& 46.1 & 71.8 & 80.7 & 198.6
& -- & -- & -- & --
& -- & -- & -- & --
& -- & -- & -- & --\\
DiscoVLA, CVPR'25
& 47.0 & 73.0 & 82.8 & 202.8
& -- & -- & -- & --
& -- & -- & -- & --
& -- & -- & -- & --\\
PIG, ICCV'25
& 48.6 & 72.8 & 81.6 & 203.0
& 31.8 & 57.3 & 68.0 & 157.1
& 64.0 & 91.5 & 96.6 & 252.1
& -- & -- & -- & --\\
\midrule
\multicolumn{17}{l}{\textit{Multimodal methods}}\\
ECLIPSE, ECCV'22
& 44.9 & 71.3 & 81.6 & 197.8
& 30.2 & 55.9 & 66.6 & 152.7
& 57.8 & 88.4 & 94.3 & 240.5
& 15.7 & -- & -- & --\\
TEFAL, ICCV'23
& 49.4 & 75.9 & 83.9 & 209.2
& 32.6 & 59.1 & 69.3 & 161.0
& 61.0 & 90.4 & 95.3 & 246.7
& 18.5 & 37.3 & 48.6 & 104.4\\
AVIGATE, CVPR'25
& 50.2 & 74.3 & 83.2 & 207.7
& 32.7 & 59.8 & 70.2 & 162.7
& 63.1 & 90.7 & 95.5 & 249.3
& 18.8 & 40.0 & 51.8 & 110.6\\
SAVE, CVPR'26
& 51.3 & 78.0 & 86.9 & 216.2
& 33.5 & 60.9 & 71.4 & 165.8
& 66.1 & 92.6 & 96.8 & 255.5
& 20.8 & 44.7 & 55.9 & 121.4\\
\midrule
\textbf{UniPRO}
& \textbf{53.9} & \textbf{79.6} & \textbf{87.4} & \textbf{220.9}
& \textbf{34.1} & \textbf{61.7} & \textbf{71.9} & \textbf{167.7}
& \textbf{67.0} & \textbf{93.1} & \textbf{96.8} & \textbf{256.9}
& \textbf{22.1} & \textbf{45.4} & \textbf{56.5} & \textbf{124.0}\\
\bottomrule
\end{tabular}
\end{table*}

\section{Main Experiments}
\label{sec:experiments}
\subsection{Datasets}
\label{sec:experimental-setup}

{\emergencystretch=3em
We evaluate text-to-video retrieval on MSR-VTT~\cite{xu2016msrvtt}, VATEX~\cite{wang2019vatex}, and Charades~\cite{sigurdsson2016charades}, using four standard benchmarks. MSR-VTT contains 10,000 videos lasting approximately 10 to 32 seconds, each paired with 20 text descriptions. The MSR-VTT-9k split uses 9,000 videos and 180,000 video and text pairs for training, with 1,000 videos for testing~\cite{yu2018jsfusion}. The MSR-VTT-7k split contains 6,513 training, 497 validation, and 2,990 test videos~\cite{xu2016msrvtt}.
VATEX contains 34,991 videos with multiple descriptions per video. Following the retrieval protocol of \citet{chen2020hgr}, we use the official training set of 25,991 videos and randomly divide the 3,000 official validation videos equally into validation and test sets. The 6,000 official test videos are excluded because their annotations are private~\cite{wang2019vatex,chen2020hgr}.
Charades contains 9,848 videos, each paired with one description. We follow its official split of 7,985 training and 1,863 test videos~\cite{sigurdsson2016charades}.\par}

\subsection{Text-to-Video Experimental Configuration}

{\emergencystretch=3em
The visual branch uses pretrained CLIP ViT-B/32 and ViT-B/16 encoders~\cite{radford2021clip}. The four-dataset comparison uses ViT-B/32; we additionally evaluate both retrieval directions on MSR-VTT-9k with ViT-B/16. The acoustic branch uses AST with a DeiT backbone pretrained on ImageNet and AudioSet~\cite{gong2021ast}. Missing or invalid audio is represented by a zero vector. Whisper large-v3 transcribes the soundtrack~\cite{radford2023whisper}, and the CLIP text encoder extracts speech semantics from the transcript~\cite{radford2021clip}. We also use a zero vector when no usable speech transcript is available, allowing incomplete examples to remain in the same training and evaluation pipeline.\par}

{\emergencystretch=3em

{\emergencystretch=3em
We train for five epochs with BertAdam, a warmup proportion of 0.1, and a 512-dimensional embedding space. Batch size is 64 for MSR-VTT-9k, VATEX, and Charades, and 96 for MSR-VTT-7k. We sample 12 frames for MSR-VTT and VATEX and 32 for Charades. Transcripts are limited to 77 tokens; acoustic and speech streams are resampled to 12 tokens. The gated fusion Transformer and query resampler each have four layers. Learning rates are \(10^{-7}\) for CLIP parameters and \(10^{-4}\) for other parameters. The contrastive margin is 0.1 for MSR-VTT and Charades and 0.05 for VATEX. These settings apply to the main text retrieval experiments; Section~\ref{sec:analysis} specifies the protocols used to study other queries and modality availability.\par}\par}

\subsection{Evaluation Metrics}

We report recall at ranks 1, 5, and 10 for both text-to-video and video-to-text retrieval. R@K measures the proportion of queries whose correct match appears among the top K candidates. We also report SumR to summarize the three recall values within each direction:
\begin{equation}
\mathrm{SumR}
=
\mathrm{R@1}
+
\mathrm{R@5}
+
\mathrm{R@10}.
\label{eq:sumr}
\end{equation}
Retrieval scores and performance differences are displayed to one decimal place. Rounding can introduce small discrepancies when subtracting or aggregating the displayed values.

\subsection{Comparison Methods}
\label{sec:sota-comparison}

We compare UniPRO with representative methods on MSR-VTT-9k, MSR-VTT-7k, VATEX, and Charades. The methods are grouped by the information used to represent candidate videos. Vision-only methods include CLIP4Clip~\cite{luo2022clip4clip}, X-Pool~\cite{gorti2022xpool}, TS2-Net~\cite{liu2022ts2net}, X-CLIP~\cite{ma2022xclip}, CLIP-ViP~\cite{xue2023clipvip}, STAN~\cite{liu2023stan}, DiCoSA~\cite{jin2023dicosa}, PromptSwitch~\cite{deng2023promptswitch}, UATVR~\cite{fang2023uatvr}, UCoFiA~\cite{wang2023ucofia}, ProST~\cite{li2023prost}, DGL~\cite{yang2024dgl}, EERCF~\cite{tian2024eercf}, TeachCLIP~\cite{tian2024teachclip}, TempMe~\cite{shen2025tempme}, DiscoVLA~\cite{shen2025discovla}, and PIG~\cite{lan2025pig}. Multimodal methods additionally use acoustic or speech information and include ECLIPSE~\cite{lin2022eclipse}, TEFAL~\cite{ibrahimi2023tefal}, AVIGATE~\cite{jeong2025avigate}, and SAVE~\cite{zhao2026save}.
\begin{table*}[!t]
\begin{minipage}[t]{0.48\textwidth}
\caption{MSR-VTT-9k V2T with ViT-B/32.}
\label{tab:v2t-b32}
\centering\small
\setlength{\tabcolsep}{2.0pt}
\begin{tabular}{@{}lrrrr@{}}
\toprule
Model & R@1 & R@5 & R@10 & SumR\\
\midrule
\multicolumn{5}{l}{\textit{Vision-only methods}}\\
CLIP4Clip~\cite{luo2022clip4clip} & 43.1 & 70.5 & 81.2 & 194.8\\
X-Pool~\cite{gorti2022xpool} & 44.4 & 73.3 & 84.0 & 201.7\\
TS2-Net~\cite{liu2022ts2net} & 45.3 & 74.1 & 83.7 & 203.1\\
X-CLIP~\cite{ma2022xclip} & 46.8 & 73.3 & 84.0 & 204.1\\
DiCoSA~\cite{jin2023dicosa} & 46.7 & 75.2 & 84.3 & 206.2\\
PromptSwitch~\cite{deng2023promptswitch} & 46.0 & 74.3 & 84.8 & 205.1\\
UATVR~\cite{fang2023uatvr} & 46.9 & 73.8 & 83.8 & 204.5\\
UCoFiA~\cite{wang2023ucofia} & 47.1 & 74.3 & 83.0 & 204.4\\
ProST~\cite{li2023prost} & 46.3 & 74.2 & 83.2 & 203.7\\
DGL~\cite{yang2024dgl} & 43.5 & 70.5 & 80.7 & 194.7\\
EERCF~\cite{tian2024eercf} & 44.7 & 74.2 & 83.9 & 202.8\\
TempMe~\cite{shen2025tempme} & 45.6 & 72.4 & 81.2 & 199.2\\
DiscoVLA~\cite{shen2025discovla} & 47.7 & 73.6 & 83.6 & 204.9\\
\midrule
\multicolumn{5}{l}{\textit{Multimodal methods}}\\
ECLIPSE~\cite{lin2022eclipse} & 44.7 & 71.3 & 82.8 & 198.8\\
TEFAL~\cite{ibrahimi2023tefal} & 47.1 & 75.1 & 84.9 & 207.1\\
AVIGATE~\cite{jeong2025avigate} & 49.7 & 75.3 & 83.7 & 208.7\\
SAVE~\cite{zhao2026save} & 48.9 & 78.0 & 86.5 & 213.4\\
\midrule
\textbf{UniPRO} & \textbf{50.5} & \textbf{79.0} & \textbf{87.7} & \textbf{217.2}\\
\bottomrule
\end{tabular}
\end{minipage}\hfill
\begin{minipage}[t]{0.50\textwidth}
\caption{MSR-VTT-9k retrieval with ViT-B/16.}
\label{tab:bidirectional-b16}
\centering\small
\setlength{\tabcolsep}{0.9pt}
\begin{tabular}{@{}lrrrrrrrr@{}}
\toprule
Model & \multicolumn{4}{c}{T2V} & \multicolumn{4}{c}{V2T}\\
\cmidrule(lr){2-5}\cmidrule(lr){6-9}
& R@1 & R@5 & R@10 & SumR & R@1 & R@5 & R@10 & SumR\\
\midrule
\multicolumn{9}{l}{\textit{Vision-only methods}}\\
CLIP4Clip~\cite{luo2022clip4clip} & 46.4 & 72.1 & 82.0 & 200.5 & 45.4 & 73.4 & 82.4 & 201.2\\
X-Pool~\cite{gorti2022xpool} & 48.2 & 73.7 & 82.6 & 204.5 & 46.4 & 73.9 & 84.1 & 204.4\\
TS2-Net~\cite{liu2022ts2net} & 49.4 & 75.6 & 85.3 & 210.3 & 46.6 & 75.9 & 84.9 & 207.4\\
X-CLIP~\cite{ma2022xclip} & 49.3 & 75.8 & 84.8 & 209.9 & 48.9 & 76.3 & 85.4 & 210.6\\
STAN~\cite{liu2023stan} & 50.0 & 75.2 & 84.1 & 209.3 & -- & -- & -- & --\\
UATVR~\cite{fang2023uatvr} & 50.8 & 76.3 & 85.5 & 212.6 & 48.1 & 76.3 & 85.4 & 209.8\\
UCoFiA~\cite{wang2023ucofia} & 49.8 & 74.6 & 83.5 & 207.9 & 49.1 & 77.0 & 83.8 & 209.9\\
ProST~\cite{li2023prost} & 49.5 & 75.0 & 84.0 & 208.5 & 48.0 & 75.9 & 85.2 & 209.1\\
DGL~\cite{yang2024dgl} & 48.3 & 71.8 & 80.6 & 200.7 & 45.7 & 74.0 & 82.9 & 202.6\\
EERCF~\cite{tian2024eercf} & 49.9 & 76.5 & 84.2 & 210.6 & 47.8 & 75.3 & 85.4 & 208.5\\
TeachCLIP~\cite{tian2024teachclip} & 48.0 & 75.9 & 83.5 & 207.4 & -- & -- & -- & --\\
TempMe~\cite{shen2025tempme} & 49.0 & 74.4 & 83.3 & 206.7 & 47.6 & 75.3 & 85.4 & 208.3\\
DiscoVLA~\cite{shen2025discovla} & 50.5 & 75.6 & 83.8 & 209.9 & 49.2 & 76.0 & 84.7 & 209.9\\
PIG~\cite{lan2025pig} & 51.2 & 75.1 & 84.5 & 210.8 & -- & -- & -- & --\\
\midrule
\multicolumn{9}{l}{\textit{Multimodal methods}}\\
TEFAL~\cite{ibrahimi2023tefal} & 49.9 & 76.2 & 84.4 & 210.5 & -- & -- & -- & --\\
AVIGATE~\cite{jeong2025avigate} & 52.1 & 76.4 & 85.2 & 213.7 & 51.2 & 77.9 & 86.2 & 215.3\\
SAVE~\cite{zhao2026save} & 54.5 & 80.3 & 87.4 & 222.2 & 51.4 & 80.1 & 87.0 & 218.5\\
\midrule
\textbf{UniPRO} & \textbf{55.0} & \textbf{81.6} & \textbf{87.5} & \textbf{224.1} & \textbf{51.9} & \textbf{80.6} & \textbf{87.5} & \textbf{220.0}\\
\bottomrule
\end{tabular}
\end{minipage}
\end{table*}

\subsection{Performance}

Table~\ref{tab:sota-comparison} shows consistent gains in aggregate recall across the four benchmarks. On MSR-VTT-9k, UniPRO achieves 53.9 R@1, 79.6 R@5, and 87.4 R@10, yielding a SumR of 220.9 and improving on SAVE by 4.7 points. SumR reaches 167.7 on MSR-VTT-7k, 256.9 on VATEX, and 124.0 on Charades, with respective gains of 1.9, 1.4, and 2.6 points. These results establish the effectiveness of the complete UniPRO pipeline across the evaluated datasets. The component ablations below assess the contribution of its supporting modules, while the subsequent analyses isolate differences between fusion schedules.

\paragraph{Reverse retrieval.}
Table~\ref{tab:v2t-b32} evaluates video queries against text candidates on MSR-VTT-9k. UniPRO achieves 50.5 R@1 and 217.2 SumR with ViT-B/32. Its R@1 exceeds AVIGATE, the strongest prior method at this cutoff, by 0.8 points. Relative to SAVE, the gains are 1.6 R@1 points and 3.8 SumR points, with improvements at all three recall cutoffs. The benefit of the learned video representation therefore extends to both the forward and reverse directions of the text--video task.

\paragraph{Visual backbone.}
Table~\ref{tab:bidirectional-b16} reports the corresponding comparison with ViT-B/16. UniPRO reaches 55.0 R@1 and 224.1 SumR for text-to-video retrieval, and 51.9 R@1 and 220.0 SumR for video-to-text retrieval. It improves on SAVE by 1.9 and 1.5 SumR points in the two directions. These results complement the four-dataset ViT-B/32 evaluation and show that the pipeline remains effective with the finer visual patch size. They assess the complete model; the analysis experiments below compare fusion schedules within their own matched configurations. Dashes in the tables denote unavailable results.

The main benchmarks measure overall retrieval quality under their standard splits. The analysis experiments use additional data constructions and query definitions to examine order sensitivity, so their scores are interpreted within each stated protocol.
\section{Ablation Experiments}
\label{sec:ablation-studies}

\paragraph{Core Component Ablation.}
Table~\ref{tab:core-component-ablation} evaluates the contributions of REM and POSR. Removing REM reduces SumR by 3.0 points for text-to-video retrieval and 0.9 points for video-to-text retrieval, supporting the value of interaction between acoustic and speech features before fusion. Removing POSR reduces SumR by 4.8 and 2.1 points and R@1 by 2.2 and 1.1 points, respectively. These changes are consistent with transcript condensation retaining useful semantics that a fixed truncation rule may discard.

{\emergencystretch=3em
Adding the original visual representation after progressive fusion reduces SumR by 2.4 and 0.4 points in the two directions. Reintroducing visual features at the output therefore does not automatically improve the learned multimodal representation; its effect depends on how that information enters the fusion process.\par}

\paragraph{Multi-Feature Soft Alignment Ablation.}
Table~\ref{tab:mfsa-ablation} examines the pairwise alignment paths. Removing alignment between vision and speech reduces SumR by 2.3 points for text-to-video retrieval and 1.0 point for video-to-text retrieval. Removing alignment between audio and speech reduces SumR by 1.9 and 1.6 points. Both paths therefore contribute beyond audiovisual alignment alone.

{\emergencystretch=3em
Retaining only audiovisual alignment produces larger decreases of 3.3 and 1.9 SumR points. Together, these results support aligning all three modality pairs before fusion. The complete set of alignment paths provides a stronger common representation for the order comparisons that follow.\par}

\begin{figure}[!b]
\centering
\includegraphics[width=\columnwidth]{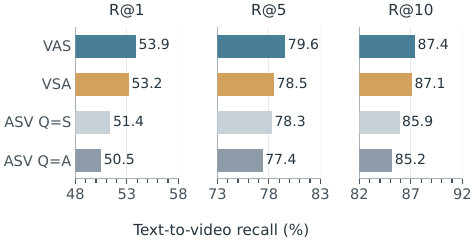}
\caption{Fusion organization on MSR-VTT-9k.}
\Description{Three panels of horizontal bars report R at 1, 5, and 10 for VAS, VSA, and two stacked ASV variants. The displayed horizontal ranges are 48--58, 73--83, and 82--92 percent, respectively.}
\label{fig:organization-bars}
\end{figure}

\begin{table*}[!t]
\begin{minipage}[t]{0.48\textwidth}
\caption{Ablation of model components.}
\label{tab:core-component-ablation}
\centering
\small
\setlength{\tabcolsep}{3pt}
\begin{tabular}{@{}lrrrr@{}}
\toprule
\textbf{Experimental Setting}
& \multicolumn{2}{c}{\textbf{Text-to-Video}}
& \multicolumn{2}{c}{\textbf{Video-to-Text}}\\
\cmidrule(lr){2-3}\cmidrule(lr){4-5}
& R@1 & SumR & R@1 & SumR\\
\midrule
w/o REM & 52.5 & 217.9 & 50.0 & 216.3\\
w/o POSR & 51.7 & 216.1 & 49.4 & 215.1\\
+ Original Visual Fusion & 52.7 & 218.5 & 50.1 & 216.8\\
\midrule
\textbf{UniPRO (Ours)} & \textbf{53.9} & \textbf{220.9} & \textbf{50.5} & \textbf{217.2}\\
\bottomrule
\end{tabular}
\end{minipage}\hfill
\begin{minipage}[t]{0.48\textwidth}
\caption{Ablation of soft alignment.}
\label{tab:mfsa-ablation}
\centering
\small
\setlength{\tabcolsep}{3pt}
\begin{tabular}{@{}lcccrrrr@{}}
\toprule
\textbf{Setting}
& \multicolumn{3}{c}{\textbf{Soft Alignment}}
& \multicolumn{2}{c}{\textbf{T2V}}
& \multicolumn{2}{c}{\textbf{V2T}}\\
\cmidrule(lr){2-4}\cmidrule(lr){5-6}\cmidrule(lr){7-8}
& V--A & V--S & A--S & R@1 & SumR & R@1 & SumR\\
\midrule
V--A Only & \ding{51} & \(\times\) & \(\times\) & 52.4 & 217.6 & 48.8 & 215.3\\
w/o V--S & \ding{51} & \(\times\) & \ding{51} & 52.4 & 218.6 & 49.4 & 216.2\\
w/o A--S & \ding{51} & \ding{51} & \(\times\) & 52.9 & 219.0 & 49.7 & 215.6\\
\midrule
\textbf{UniPRO (Ours)} & \ding{51} & \ding{51} & \ding{51} & \textbf{53.9} & \textbf{220.9} & \textbf{50.5} & \textbf{217.2}\\
\bottomrule
\end{tabular}
\end{minipage}
\end{table*}

\section{Analysis Experiments}
\label{sec:analysis}
We study fusion organization, modality availability, and query compatibility. CM-Pool pools complete samples from MSR-VTT, VATEX, and Charades for text retrieval controls. MSR-Avail groups the two MSR-VTT test splits by natural modality availability. MSR-Complete retains complete MSR-VTT samples for audio retrieval; MSR-SDrop introduces speech missingness in both its training and test sets. MSR-Frame uses the full MSR-VTT-9k split for image queries. Appendix~\ref{app:analysis-data} details construction and settings. Plots summarize the key comparisons, and Table~\ref{tab:text-input-removal} reports the inference-only control. Appendix~\ref{app:additional-analyses} provides the complete numerical results underlying the plots.

\subsection{Fusion Organization and Order}
\label{sec:order-analysis}
We train and test on the full MSR-VTT-9k split, including naturally incomplete examples, with matched encoders, components, and training settings. Stacked ASV first combines audio and speech, with either S or A as its attention query Q, then introduces vision. VAS and VSA instead update the visual stream one modality at a time.

Figure~\ref{fig:organization-bars} shows that both one-by-one schedules outperform the stacked variants at all three recall cutoffs. VAS also improves R@1 over VSA by 0.7 percentage points. Fusion organization and the order within that organization therefore both matter. We next examine the roles of learning difficulty and information preservation.

\subsection{Principle I}
The first principle concerns learning difficulty under unequal input availability. All videos provide V, fewer provide usable A, and fewer still provide S. In VSA, the second stage combines audio with V when speech is absent and with VS when it is present. In VAS, an informative final S interaction receives VA because usable speech also implies available audio. This motivates a path with simpler effective input patterns, although pattern counts alone do not measure optimization difficulty.

\begin{figure}[!htbp]
\centering
\includegraphics[width=\columnwidth]{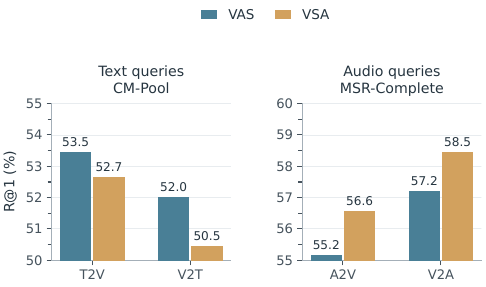}
\caption{R@1 with complete inputs: text on CM-Pool and audio on MSR-Complete.}
\Description{Two panels compare VAS and VSA in both retrieval directions. Text and audio experiments use different data constructions.}
\label{fig:query-order-bars}
\end{figure}

To control natural missingness, we select complete training and evaluation samples separately from MSR-VTT, VATEX, and Charades and concatenate the corresponding portions into CM-Pool. Figure~\ref{fig:query-order-bars} shows that VAS remains better: T2V R@1 is 53.5 versus 52.7, and V2T R@1 is 52.0 versus 50.5. Missing inputs therefore cannot fully explain order sensitivity. The frozen-model control below tests whether this learned difference persists when modalities are removed only at inference.

\paragraph{Input removal after complete training.}
\label{sec:text-input-removal}

We next freeze the two text retrieval checkpoints trained on complete CM-Pool inputs. The same complete evaluation pool is tested with V, VA, all three inputs, and VS by zeroing the removed streams. Queries and candidate identities are unchanged, and neither model is retrained. This separates removing information at inference from learning a representation with that information present throughout training.

VAS and VSA are trained with matched settings apart from the fusion order. Table~\ref{tab:text-input-removal} shows that VAS maintains higher T2V R@1 across all four test input conditions. Its R@1 is 47.7 versus 45.6 with V alone and 50.6 versus 49.7 with VA. Thus, the performance ranking associated with the training order persists when the available test modalities change.
\begin{table}[!htbp]
\caption{Text-to-video R@1 with test-time input removal.}
\label{tab:text-input-removal}
\centering
\small
\setlength{\tabcolsep}{2.4pt}
\begin{tabular}{@{}lcccr@{}}
\toprule
Order & \multicolumn{3}{c}{Input} & T2V\\
\cmidrule(lr){2-4}\cmidrule(lr){5-5}
& V & A & S & R@1\\
\midrule
VAS & \ding{51} & \(\times\) & \(\times\) & 47.7\\
VAS & \ding{51} & \ding{51} & \(\times\) & 50.6\\
VAS & \ding{51} & \ding{51} & \ding{51} & 53.5\\
VAS & \ding{51} & \(\times\) & \ding{51} & 49.7\\
VSA & \ding{51} & \(\times\) & \(\times\) & 45.6\\
VSA & \ding{51} & \ding{51} & \(\times\) & 49.7\\
VSA & \ding{51} & \ding{51} & \ding{51} & 52.7\\
VSA & \ding{51} & \(\times\) & \ding{51} & 49.5\\
\bottomrule
\end{tabular}
\end{table}

\subsection{Principle II}
\label{sec:information-preservation}
VAS still outperforms VSA when all three modalities are present during training and evaluation. Thus, missing modalities cannot fully explain why one order works better. We therefore seek a second principle to explain the remaining difference.

To identify this second principle, we compare retrieval performance across the three MSR-Avail groups in MSR-VTT-9k and MSR-VTT-7k. For each split, we use the same trained VAS model and full video candidate pool. Table~\ref{tab:natural-groups} shows that the V+A+S group achieves the highest R@1 and R@5 on both splits, outperforming the V+A and V-only groups.

In VAS, S enters last. S represents the speech transcript, which is text like the retrieval query. This suggests a hypothesis: \emph{introducing the modality that matches the query last may help retain information useful for retrieval.} We adopt this as Principle II. Since the groups contain different videos, this observation only suggests the hypothesis. We therefore test it in another retrieval task.
\begin{table}[!htbp]
\caption{Text-to-video retrieval by modality availability.}
\label{tab:natural-groups}
\centering
\small
\setlength{\tabcolsep}{2.4pt}
\begin{tabular}{@{}cccrrrr@{}}
\toprule
V & A & S & Videos & R@1 & R@5 & R@10\\
\midrule
\multicolumn{7}{l}{\textit{MSR-VTT-9k}}\\
\midrule
\ding{51} & \ding{51} & \ding{51} & 638 & 55.4 & 81.6 & 88.2\\
\ding{51} & \ding{51} & \(\times\) & 307 & 53.5 & 77.2 & 87.9\\
\ding{51} & \(\times\) & \(\times\) & 55 & 39.2 & 69.7 & 75.6\\
\multicolumn{3}{l}{All} & 1000 & 53.9 & 79.6 & 87.4\\
\midrule
\multicolumn{7}{l}{\textit{MSR-VTT-7k}}\\
\midrule
\ding{51} & \ding{51} & \ding{51} & 1872 & 34.9 & 62.3 & 72.1\\
\ding{51} & \ding{51} & \(\times\) & 963 & 33.9 & 62.1 & 72.7\\
\ding{51} & \(\times\) & \(\times\) & 155 & 26.4 & 52.7 & 65.1\\
\multicolumn{3}{l}{All} & 2990 & 34.1 & 61.7 & 71.9\\
\bottomrule
\end{tabular}
\end{table}

\label{sec:audio-query-transfer}
For audio queries, this hypothesis predicts that VSA will perform better than VAS, because VSA introduces audio last. We test this prediction on MSR-Complete, using complete MSR-VTT samples for both training and evaluation. Each audio query is a fixed contiguous crop lasting 1/20 of its source soundtrack. The 638 video candidates retain V, the full soundtrack A containing the query crop, and S. We evaluate retrieval in both directions; Appendix~\ref{app:msr-complete} details the protocol.

Figure~\ref{fig:query-order-bars} supports the prediction: VSA improves A2V R@1 from 55.2 to 56.6 and V2A R@1 from 57.2 to 58.5. Thus, the preferred order places S last for text queries and A last for audio queries. This agreement supports Principle II at R@1. It does not hold at every cutoff: A2V R@5 is tied, and VAS has higher R@10. The text and audio experiments use different datasets, so their comparison does not isolate the effect of query modality alone.

\subsection{Balancing the Principles}
\label{sec:speech-missingness}
On MSR-Complete, we remove S from the same proportion of training and test samples, retaining V and A. At 10\%, each split contains 90\% V+A+S and 10\% V+A samples. We train new VAS and VSA models at every rate from 0\% to 100\% under matched settings. Audio queries and test candidates stay fixed. Appendix~\ref{app:msr-sdrop} gives the configuration.

With partial S missingness, VSA's audio fusion stage must handle both V and VS. For VAS, the final stage still combines VA with S whenever S is available. VSA therefore faces more varied inputs as considered in Principle I, while still following Principle II by placing audio last.

\begin{figure}[!htbp]
\centering
\includegraphics[width=\columnwidth]{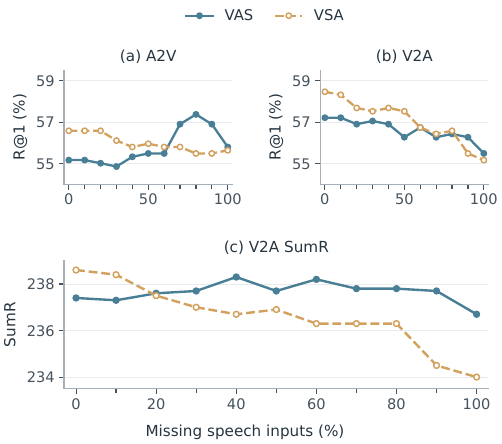}
\caption{R@1 and SumR under speech missingness.}
\Description{Two upper panels compare VAS and VSA R at 1 for audio-to-video and video-to-audio retrieval. A full-width lower panel shows V2A SumR, the sum of the three displayed V2A recalls, at eleven speech missingness rates, with separate training of each order at each rate and matched missingness proportions in training and testing.}
\label{fig:speech-missingness-curves}
\end{figure}

Figure~\ref{fig:speech-missingness-curves} shows reversals in both R@1 and SumR. VSA has higher A2V R@1 through 60\% missingness, whereas VAS is higher at 70\%--90\%. With S fully absent, the scores are similar, at 55.8 versus 55.6. V2A SumR favors VSA at 0\%--10\%, then VAS from 20\% onward. The crossover points depend on the retrieval direction and metric. These results support balancing the two principles as modality availability changes. Exact values appear in Appendix~\ref{app:missingness-results}, Tables~\ref{tab:missingness} and~\ref{tab:missingness-reverse}.

\subsection{Predicting an Order for Image Queries}
\label{sec:image-query-transfer}
We transfer the insights to MSR-Frame, using the full MSR-VTT-9k split of 9,000 training and 1,000 test videos. One random source frame serves as each image query and is fixed across schedules. Candidate videos retain their naturally available V, full A, and S. This evaluates source-frame matching rather than retrieval from unrelated external images.

The principles predict VASV: follow the visual, acoustic, and speech path, then revisit the original visual tokens last. VSAV tests the preceding A/S order; VVAS and VAVS test earlier positions for repeated V. All four repeated-V schedules have three independently parameterized gated stages, controlling for extra depth; VAS is the original two-stage reference.

\begin{figure}[!htbp]
\centering
\includegraphics[width=\columnwidth]{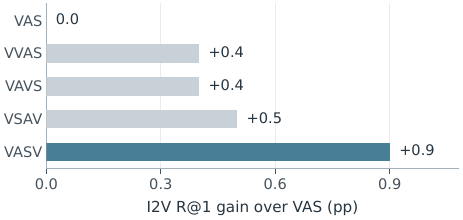}
\caption{Image retrieval gain over VAS, whose R@1 is 97.0\%.}
\Description{Bars show percentage-point R at 1 gains relative to VAS: 0.4 for VVAS and VAVS, 0.5 for VSAV, and 0.9 for VASV.}
\label{fig:image-order-bars}
\end{figure}

Figure~\ref{fig:image-order-bars} supports the prediction. VASV achieves I2V R@1 of 97.9, exceeding VAS by 0.9 points, VSAV by 0.4, and VVAS/VAVS by 0.5. Its R@5 and R@10 are also at least as high as the tested alternatives. This supports transfer within the evaluated protocol, while broader robustness remains to be established. Complete metrics and baseline comparisons are in Appendix~\ref{app:image-results}.

\section{Conclusion}
UniPRO makes fusion order an explicit choice within a common framework for progressive multimodal representation learning. Benchmark comparisons and component ablations establish the effectiveness of the complete pipeline. The evaluated one-by-one schedules outperform stacked fusion, while complete-input controls show that missingness alone cannot explain the remaining order differences. Learning difficulty and preservation of information relevant to the query offer complementary guidance. Reversals in both R@1 and SumR under speech missingness in training and testing motivate balancing these considerations, and their transfer predicts the best evaluated image retrieval order. Repeated-run evaluation and broader query protocols can further test the generality of these empirical principles. Fusion order should therefore reflect both the information available in a video and the information needed by its query.

\FloatBarrier
\label{sec:main-end}
\clearpage
\phantomsection\label{sec:references-start}
\begin{figure*}[!t]
\begin{minipage}[t]{0.48\textwidth}
\vspace{0pt}
\centering
\includegraphics[width=\linewidth]{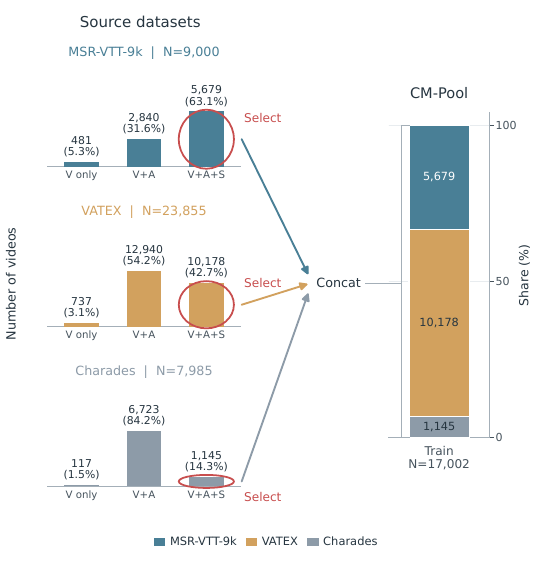}
\caption{CM-Pool training construction. Red ellipses mark selected complete-modality groups.}
\Description{For each of three training sources, bars ordered V only, V+A, and V+A+S show counts and percentages. Red ellipses select the rightmost complete groups: 5,679 MSR-VTT-9k, 10,178 VATEX, and 1,145 Charades videos. Three arrows connect the selected groups to the pool under a Concat label, yielding 17,002 training videos.}
\label{fig:cm-pool-construction}
\end{minipage}\hfill
\begin{minipage}[t]{0.48\textwidth}
\vspace{0pt}
\centering
\includegraphics[width=\linewidth]{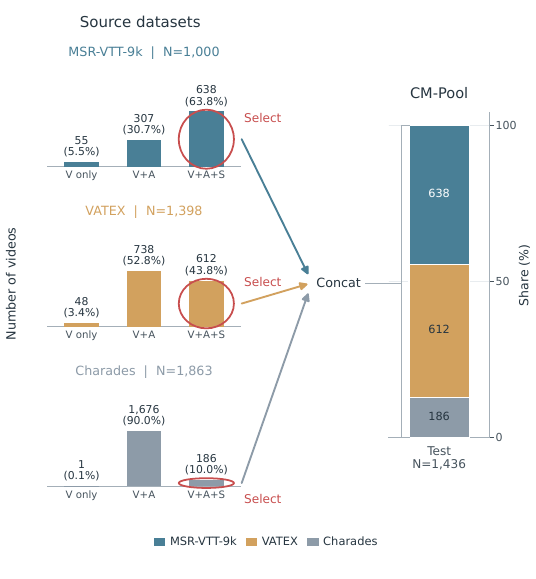}
\caption{CM-Pool test construction, using the same selection and concatenation procedure.}
\Description{Three test sources show modality groups ordered V only, V+A, and V+A+S, with counts and percentages. Red ellipses select complete groups of 638 MSR-VTT-9k, 612 VATEX, and 186 Charades videos, with a Select label beside each ellipse and three arrows pointing to the Concat pool of 1,436 test videos.}
\label{fig:cm-pool-test-construction}
\end{minipage}
\end{figure*}
\bibliographystyle{ACM-Reference-Format}
\bibliography{UniPRO}
\appendix
\section{Construction of the Analysis Datasets}
\label{app:analysis-data}
The following protocols distinguish natural availability from controlled removal. V denotes video frames, A the soundtrack, and S usable ASR-derived speech. All schedules within a comparison share source examples, queries, and candidates.

\subsection{CM-Pool}
\label{app:cm-pool}
Figures~\ref{fig:cm-pool-construction} and~\ref{fig:cm-pool-test-construction} summarize the construction of the CM-Pool training and test splits. We independently filter the training and test splits of MSR-VTT-9k, VATEX, and Charades, retaining videos with decodable visual content, valid nonzero audio, usable ASR transcripts, and associated text descriptions. The source splits contain 9,000/1,000 training/test videos for MSR-VTT-9k, 23,896/1,398 for locally available VATEX, and 7,985/1,863 for Charades. Filtering retains 5,679/638, 10,178/612, and 1,145/186 videos, respectively. We concatenate the retained training portions and test portions separately, yielding 17,002 training and 1,436 test videos, or 18,438 videos in total. Training and test identities remain disjoint, and dataset-specific ID prefixes prevent naming collisions. The separate VATEX validation split is excluded from this pool. The source modality distributions exclude 41 VATEX training videos with unavailable visual content; percentages are normalized over the remaining valid videos within each source split.

To avoid unequal caption multiplicities across sources, each training video contributes one deterministically selected text query per epoch. Each test video has one fixed text query, yielding 1,436 queries and 1,436 video candidates. VAS and VSA share the same videos and query-selection protocol.

\subsection{MSR-Avail}
\label{app:msr-avail}
\begin{figure}[!htbp]
\centering
\includegraphics[width=\columnwidth]{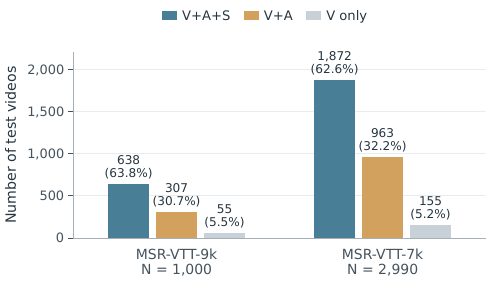}
\caption{Natural modality availability in MSR-Avail.}
\Description{Grouped bars show 638 complete, 307 audiovisual without speech, and 55 vision-only videos for MSR-VTT-9k; corresponding counts for MSR-VTT-7k are 1,872, 963, and 155. Percentages are 63.8, 30.7, and 5.5 percent for the former and 62.6, 32.2, and 5.2 percent for the latter.}
\label{fig:msr-availability-counts}
\end{figure}

MSR-Avail partitions the test videos of MSR-VTT-9k and MSR-VTT-7k into three mutually exclusive groups according to their naturally available inputs: V+A+S, V+A without S, and V only. Figure~\ref{fig:msr-availability-counts} shows their sizes and within-benchmark proportions. MSR-VTT-9k contains 638, 307, and 55 videos in the three groups, respectively, totaling 1,000 videos; MSR-VTT-7k contains 1,872, 963, and 155, totaling 2,990 videos. Complete videos account for 63.8\% and 62.6\% of the respective test sets, while vision-only videos account for 5.5\% and 5.2\%.

\subsection{MSR-Complete and MSR-SDrop}
\label{app:msr-complete}
MSR-Complete filters MSR-VTT training and evaluation samples for valid V, A, and S, retaining 5,679 training videos and 638 evaluation videos. For a soundtrack of length L, a random contiguous crop of length L/20 becomes the audio query. The crop is fixed across orders and removal conditions. Its positive candidate retains the full soundtrack containing the crop, video frames, and speech. A2V ranks the 638 videos; V2A ranks their associated audio crops.

\paragraph{Training.}
The complete-input VAS and VSA models use CLIP ViT-B/32, AST, POSR, REM, and three equally weighted alignment paths. Both use five epochs, global batch size 64, learning rate \(10^{-4}\), and seed 42. For MSR-SDrop, we repeat these training settings separately for each fusion order and each missing rate, with one run per configuration. A model is evaluated at the missing rate used for its training; the complete-input checkpoints are not reused across rates.

\paragraph{Speech removal.}
\label{app:msr-sdrop}
MSR-SDrop starts from MSR-Complete's 5,679 training videos and 638 test videos. At each rate from 0\% to 100\% in steps of 10 percentage points, the same proportion of samples in each split loses S, while V and A remain available. For example, at 10\%, each split contains 90\% complete samples and 10\% with only V+A. S token sequences and attention/segment masks are zeroed for the selected samples before training and evaluation. No samples are removed from the pools.

For each rate, VAS and VSA are trained separately on the corresponding incomplete training set and evaluated on the corresponding incomplete test set. Audio query crops and candidate identities stay fixed across rates. The test masks affect nested subsets of 0, 64, 128, 191, 255, 319, 383, 447, 510, 574, and 638 videos. The affected test videos are shared across orders. Encoders, REM, and fusion stages remain active. At 100\% missingness, both orders receive only V and A, but their parameters were optimized separately; small score differences from single runs should not be overinterpreted.

\subsection{MSR-Frame}
\label{app:msr-frame}
We use the full MSR-VTT-9k split of 9,000 training and 1,000 test videos. One random source frame per video supplies its image query, fixed across orders and encoded independently. Candidates retain their natural A and S availability rather than being filtered to complete samples. Frames are extracted at 1 fps, uniformly selecting at most 12, with padding for shorter sequences and \(224\times224\) inputs. I2V ranks 1,000 videos; V2I ranks their associated image queries. Queries come from their positive videos, so this measures source-frame matching. All schedules use CLIP ViT-B/32, AST, POSR, REM, and the three alignment paths, trained for five epochs with global batch size 64 and learning rate \(10^{-4}\). Repeated-V schedules have three learned gated stages and reuse the original visual tokens with frame positions added once; VAS has two stages.

\paragraph{Additional benchmark settings.}
The inherited text retrieval configuration additionally uses intra-negative rate \(\beta=0.2\), local aggregation temperature \(\tau=1\), and MFSA scale \(\alpha=5\), with equal pairwise weights of \(1/3\). Other main benchmark settings are given in Section~\ref{sec:experimental-setup}.

\section{Further Analysis Results}
\label{app:additional-analyses}
The tables below retain complete numerical results for the plotted order comparisons. Natural modality groups and the fixed-checkpoint input-removal control are reported in Tables~\ref{tab:natural-groups} and~\ref{tab:text-input-removal} in the main paper.

\subsection{Fusion Organization and Complete Inputs}
Table~\ref{tab:orders-new} complements Figure~\ref{fig:organization-bars}. Tables~\ref{tab:complete-modalities} and~\ref{tab:audio-orders} give both retrieval directions for CM-Pool and MSR-Complete. The two constructions have different source datasets. The natural-group comparison is reported in Table~\ref{tab:natural-groups} in the main paper.

\begin{table}[!htbp]
\caption{Fusion organization and order on MSR-VTT-9k.}
\label{tab:orders-new}
\centering
\small
\setlength{\tabcolsep}{2.4pt}
\begin{tabular}{lrrr}
\toprule
Schedule & R@1 & R@5 & R@10\\
\midrule
VAS & 53.9 & 79.6 & 87.4\\
VSA & 53.2 & 78.5 & 87.1\\
ASV (Q=S) & 51.4 & 78.3 & 85.9\\
ASV (Q=A) & 50.5 & 77.4 & 85.2\\
\bottomrule
\end{tabular}
\end{table}

\begin{table}[!htbp]
\caption{Fusion order on CM-Pool.}
\label{tab:complete-modalities}
\centering
\small
\setlength{\tabcolsep}{2.4pt}
\begin{tabular}{lrrrrrr}
\toprule
\multicolumn{1}{c}{Schedule} & \multicolumn{3}{c}{T2V} & \multicolumn{3}{c}{V2T}\\
\cmidrule(lr){2-4}\cmidrule(lr){5-7}
& R@1 & R@5 & R@10 & R@1 & R@5 & R@10\\
\midrule
VAS & 53.5 & 82.5 & 89.8 & 52.0 & 81.1 & 89.7\\
VSA & 52.7 & 81.2 & 88.9 & 50.5 & 80.2 & 88.7\\
\bottomrule
\end{tabular}
\end{table}

\begin{table}[!htbp]
\caption{Audio--video retrieval on MSR-Complete.}
\label{tab:audio-orders}
\centering
\small
\setlength{\tabcolsep}{2.4pt}
\begin{tabular}{lrrrrrr}
\toprule
\multicolumn{1}{c}{Schedule} & \multicolumn{3}{c}{A2V} & \multicolumn{3}{c}{V2A}\\
\cmidrule(lr){2-4}\cmidrule(lr){5-7}
& R@1 & R@5 & R@10 & R@1 & R@5 & R@10\\
\midrule
VAS & 55.2 & 87.2 & 94.8 & 57.2 & 86.5 & 93.7\\
VSA & 56.6 & 87.2 & 93.1 & 58.5 & 86.4 & 93.7\\
\bottomrule
\end{tabular}
\end{table}

\subsection{Controlled Speech Removal}
\label{app:missingness-results}
Tables~\ref{tab:missingness} and~\ref{tab:missingness-reverse} give the values underlying Figure~\ref{fig:speech-missingness-curves}. Each row uses VAS and VSA models trained separately at that missing rate, with the same missing proportion in training and testing. Audio crops and candidate identities stay fixed across rates; Appendix~\ref{app:msr-sdrop} gives the protocol.
\begin{table}[!htbp]
\caption{Audio-to-video retrieval with S missingness.}
\label{tab:missingness}
\centering
\small
\setlength{\tabcolsep}{2.4pt}
\begin{tabular}{rrrrrrrr}
\toprule
S (\%) & Videos & \multicolumn{3}{c}{VAS} & \multicolumn{3}{c}{VSA}\\
\cmidrule(lr){3-5}\cmidrule(lr){6-8}
& & R@1 & R@5 & R@10 & R@1 & R@5 & R@10\\
\midrule
0 & 0 & 55.2 & 87.2 & 94.8 & 56.6 & 87.2 & 93.1\\
10 & 64 & 55.2 & 86.7 & 94.2 & 56.6 & 86.8 & 93.1\\
20 & 128 & 55.0 & 86.7 & 94.0 & 56.6 & 87.0 & 93.3\\
30 & 191 & 54.9 & 87.0 & 94.5 & 56.1 & 86.8 & 93.3\\
40 & 255 & 55.3 & 86.7 & 94.5 & 55.8 & 86.5 & 93.1\\
50 & 319 & 55.5 & 86.2 & 94.2 & 56.0 & 86.5 & 93.0\\
60 & 383 & 55.5 & 86.4 & 94.0 & 55.8 & 85.9 & 93.1\\
70 & 447 & 56.9 & 86.5 & 94.0 & 55.8 & 86.5 & 93.3\\
80 & 510 & 57.4 & 86.5 & 93.7 & 55.5 & 86.8 & 93.3\\
90 & 574 & 56.9 & 86.2 & 93.7 & 55.5 & 86.5 & 93.3\\
100 & 638 & 55.8 & 85.9 & 93.6 & 55.6 & 86.4 & 93.1\\
\bottomrule
\end{tabular}
\end{table}

\begin{table}[!htbp]
\caption{Video-to-audio retrieval with S missingness.}
\label{tab:missingness-reverse}
\centering
\small
\setlength{\tabcolsep}{2.4pt}
\begin{tabular}{rrrrrrrr}
\toprule
S (\%) & Videos & \multicolumn{3}{c}{VAS} & \multicolumn{3}{c}{VSA}\\
\cmidrule(lr){3-5}\cmidrule(lr){6-8}
& & R@1 & R@5 & R@10 & R@1 & R@5 & R@10\\
\midrule
0 & 0 & 57.2 & 86.5 & 93.7 & 58.5 & 86.4 & 93.7\\
10 & 64 & 57.2 & 86.5 & 93.6 & 58.3 & 86.4 & 93.7\\
20 & 128 & 56.9 & 87.0 & 93.7 & 57.7 & 86.1 & 93.7\\
30 & 191 & 57.1 & 87.0 & 93.6 & 57.5 & 85.9 & 93.6\\
40 & 255 & 56.9 & 87.5 & 93.9 & 57.7 & 85.4 & 93.6\\
50 & 319 & 56.3 & 87.5 & 93.9 & 57.5 & 85.7 & 93.7\\
60 & 383 & 56.7 & 87.5 & 94.0 & 56.7 & 85.7 & 93.9\\
70 & 447 & 56.3 & 87.5 & 94.0 & 56.4 & 85.7 & 94.2\\
80 & 510 & 56.4 & 87.2 & 94.2 & 56.6 & 85.7 & 94.0\\
90 & 574 & 56.3 & 87.2 & 94.2 & 55.5 & 85.1 & 93.9\\
100 & 638 & 55.5 & 87.0 & 94.2 & 55.2 & 85.1 & 93.7\\
\bottomrule
\end{tabular}
\end{table}

\FloatBarrier
\subsection{Image Retrieval}
\label{app:image-results}
Table~\ref{tab:image-orders} reports all available image retrieval metrics and the CLIP4Clip~\cite{luo2022clip4clip} and ViCLIP~\cite{wang2024internvid} baselines. VASV exceeds their listed I2V R@1 values by 5.5 and 13.3 points. Unavailable reverse scores remain unspecified.
\par\noindent\begin{minipage}{\columnwidth}
\captionof{table}{Image--video retrieval on MSR-Frame.}
\label{tab:image-orders}
\centering
\small
\setlength{\tabcolsep}{2.4pt}
\begin{tabular}{lrrrrrr}
\toprule
\multicolumn{1}{c}{Method} & \multicolumn{3}{c}{I2V} & \multicolumn{3}{c}{V2I}\\
\cmidrule(lr){2-4}\cmidrule(lr){5-7}
& R@1 & R@5 & R@10 & R@1 & R@5 & R@10\\
\midrule
\shortstack[l]{CLIP4Clip,\\Neurocomputing'22} & 92.4 & -- & -- & -- & -- & --\\
\shortstack[l]{ViCLIP, ICLR'24} & 84.6 & -- & -- & -- & -- & --\\
VAS & 97.0 & 98.8 & 99.3 & 95.7 & 99.0 & 99.4\\
VVAS & 97.4 & 99.4 & 99.6 & -- & -- & --\\
VAVS & 97.4 & 99.4 & 99.4 & -- & -- & --\\
VSAV & 97.5 & 99.1 & 99.4 & 96.0 & 99.4 & 99.5\\
VASV & 97.9 & 99.4 & 99.6 & -- & -- & --\\
\bottomrule
\end{tabular}
\end{minipage}

\FloatBarrier
\end{document}